\pdfoutput=1

\documentclass[11pt]{article}

\usepackage[preprint]{acl_latex/acl}

\usepackage{times}
\usepackage{latexsym}

\usepackage[T1]{fontenc}

\usepackage[utf8]{inputenc}

\usepackage{microtype}

\usepackage{inconsolata}

\usepackage{graphicx}

\usepackage{amsmath}
\usepackage{amssymb}
\usepackage{mathtools}
\usepackage{amsthm}

\usepackage{microtype}
\usepackage{graphicx}
\usepackage{subfigure}
\usepackage{booktabs} 
\usepackage{soul,enumitem}
\usepackage{algorithm}
\usepackage{algpseudocode}
\usepackage{caption}
\usepackage{subcaption}
\usepackage{multirow}

\makeatletter
\newcommand{\IfPreprintOrFinal}[1]{%
  \@ifpackagewith{neurips_2025}{final}{#1}{%
    \@ifpackagewith{neurips_2025}{preprint}{#1}{%
        \newpage
    }%
  }%
}
\makeatother

\title{ClusterFewshot: Improving Few-shot Optimization for LLMs workflow}

\author{Omri Bar Haim ~~~~~ ~~~~~ Shahar Katz ~~~~~  ~~~~~ Lior Wolf\\
Blavatnik School of Computer Science, Tel Aviv University\\
\small{\texttt{\{omribarhaim@mail,shaharkatz3@mail,wolf@cs\}.tau.ac.il}}
}

\begin{document}
\maketitle

\begin{abstract}
The performance of large language model (LLM) workflows often depends on selecting a small set of in-context demonstrations to guide model behavior on new tasks. {Recent methods improve} this process by augmenting prompts with successful reasoning paths. {However, their demonstration} selection relies on random sampling or metric-based rankings, overlooking the semantic structure of the task. We propose \textbf{ClusterFewshot}, a strategy that combines semantic structuring with utility-aware scoring to construct representative and effective few-shot demonstration sets. Evaluated within DSPy-based pipelines, ClusterFewshot substantially reduces optimization cost across multiple benchmarks, while consistently improving accuracy relative to prior bootstrap-based methods in both standalone prompt tuning and hybrid prompt-weight optimization. Our code is available at \url{https://github.com/omrirh/clusterfewshot}.
\end{abstract}

\section{Introduction and Related Work}
Recent work has highlighted the importance of semantic structure in LLM few-shot prompt
construction, suggesting that semantically diverse and representative demonstrations can substantially improve in-context learning performance
\cite{levy2023diversedemonstrationsimproveincontext, c-etal-2024-improving}.
These findings suggest that demonstration selection extends beyond surface-level heuristic procedures,
and can be naturally viewed as a structured optimization problem over the semantic space of examples.

Today, prompt optimization and demonstration selection are often implemented through high-level toolkits that organize complex reasoning pipelines, such as LangChain and LlamaIndex \cite{Chase_LangChain_2022, Liu_LlamaIndex_2022}. These frameworks often rely on manually constructed prompt templates, including few-shot prompting \cite{brown2020language, yang2024llmoptim, opsahl-ong-etal-2024-multistage, yao-etal-2024-samples, chen2023demonstrationsneedincontextlearning, kim2022selfgeneratedincontextlearningleveraging, do2024promptoptimizationadversarialincontext}.

\begin{figure}[h]
    \centering
    \includegraphics[width=0.73\linewidth]{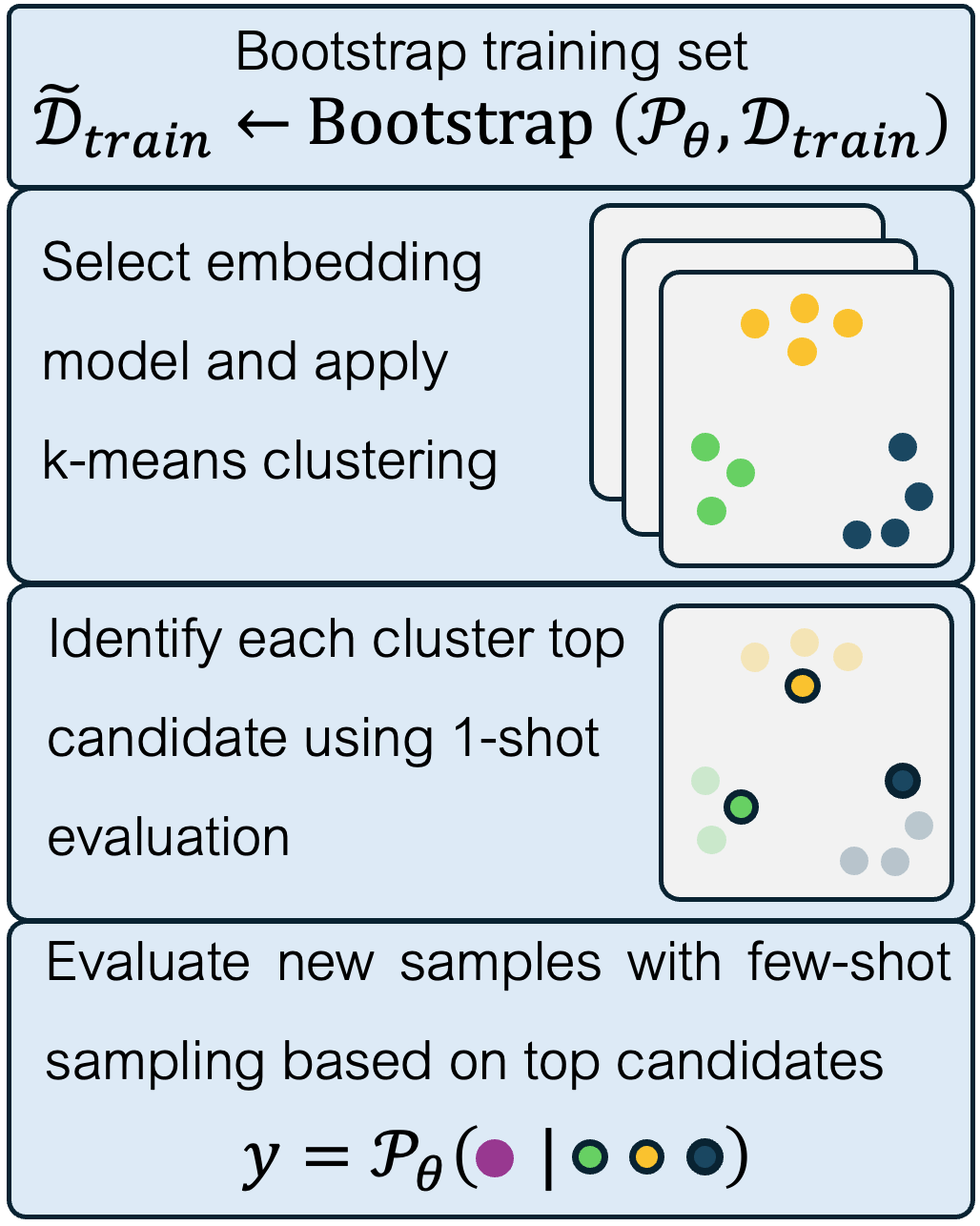}
    \caption{
    An illustration depicting ClusterFewshot approach for bootstrapped demonstrations.}
    \label{fig:overview}
\end{figure}

\citet{khattab2024dspy} introduced DSPy, a declarative framework that enables modular construction and optimization of LLM pipelines. DSPy supports retrieval-augmented generation (RAG) and enables prompt optimization through \emph{bootstrapping}, which extracts reasoning paths from solvable examples to construct effective in-context demonstrations \cite{khattab2024dspy, opsahl-ong-etal-2024-multistage}.
Subsequently, \citet{soylu-etal-2024-bettertogether} introduced \textbf{BetterTogether}, a hybrid optimization strategy that interleaves prompt-level adaptation with parameter-efficient fine-tuning.
While \citet{opsahl-ong-etal-2024-multistage, soylu-etal-2024-bettertogether} were able to achieve strong results on diverse tasks, including GSM8K \cite{cobbe2021gsm8k}, HotPotQA \cite{yang2018hotpotqa} and Iris classification \cite{fisher1936use}, 
their approach to few-shot demonstration selection typically relies on random search or metric-based ranking, lacking the semantic information that other work identifies \cite{levy2023diversedemonstrationsimproveincontext, c-etal-2024-improving}.
Such semantic insights have advanced compositional parsing \cite{levy2023diversedemonstrationsimproveincontext} and clinical NER \cite{c-etal-2024-improving}, yet remain siloed from automated few-shot optimization workflows.
{This limitation extends beyond standard LLM settings to agentic
environments such as ReAct~\cite{yao2023reactsynergizingreasoningacting},
where the target model is augmented with external tools.}

In this work, we address the lack of semantically driven selection in existing bootstrapping approaches by introducing \textbf{ClusterFewshot}, a new semantically informed prompt optimizer. Our proposed method (i) clusters embeddings of training examples to promote diverse, representative sampling, and (ii) scores candidate demonstrations using feedback from one-shot evaluation on a held-out validation subset, as illustrated in \autoref{fig:overview}, thereby guiding few-shot construction beyond random or heuristic ranking.

We focus our evaluation on bootstrap-based optimizers that operate within the
\textbf{BetterTogether} framework.
In particular, we compare with \textbf{BootstrapFewShotRS (BFRS)} \cite{khattab2024dspy},
a few-shot selection method relying on random search, and
\textbf{MIPROv2} \cite{opsahl-ong-etal-2024-optimizing}, which performs joint optimization of instructions and demonstrations via Bayesian search.
This choice enables the study of search-driven optimization over prompt components under a shared bootstrap demonstration construction paradigm.

ClusterFewshot applies to both standalone prompt optimization and hybrid tuning pipelines such as BetterTogether. Across multiple tasks and models, it consistently improves over previous approaches, including combined prompt and parameter-efficient fine-tuning pipelines.

Our main contributions are:
(i) we identify semantically informed example selection as a key limitation in existing procedures, (ii) we propose ClusterFewshot, which combines semantic clustering and evaluation-driven scoring to guide few-shot demonstration selection, and (iii) we empirically show that ClusterFewshot substantially reduces optimization {cost} across GSM8K, HotPotQA, and Iris benchmarks while preserving competitive, and often improved, mean accuracy in both prompt-only and hybrid prompt-weight optimization. {In addition, (iv) we examine the flexibility of ClusterFewshot in ReAct-based agentic setups, and (v) we evaluate its robustness across different model sizes, model families, and diverse cluster sampling approaches.}

\section{Background}
\label{Background}
We consider LLM-based workflows such as multi-hop reasoning in retrieval-augmented generation (RAG) tasks. Each task is defined by a base LLM and a labeled dataset, split into training \(\mathcal{D}_{\mathrm{train}} = \{(x_i, y_i)\}_{i=1}^N\), validation \(\mathcal{D}_{\mathrm{val}} = \{(x_j, y_j)\}_{j=1}^M\), and test \(\mathcal{D}_{\mathrm{test}}\) sets. {Each workflow is represented as a graph of modules \(\mathcal{P}_\theta\), which is} optimized (``compiled’’) to maximize performance on a held-out validation set $\mathcal{D}_{\mathrm{val}}$, through mechanisms such as fine-tuning, bootstrapped demonstration selection, and instruction refinement.
We consider several \emph{bootstrap-based} techniques, employed for prompt component generation and fine-tuning workflows.

{For in-context demonstration selection, \textbf{BootstrapFewShotRS (BFRS)} \cite{khattab2024dspy} is a} random search implementation that selects \(k \in \mathbb{N}_0\) labeled or solvable examples from \(\mathcal{D}_{\mathrm{train}}\), {which are then} used to generate reasoning traces via prompting techniques such as Chain-of-Thought (CoT) \cite{wei2022chain}. These traces are prepended to validation examples and evaluated empirically on \(\mathcal{D}_{\mathrm{val}}\). The best-performing examples and their traces are concatenated to form the final few-shot prompt.

{More recently, \textbf{MIPROv2} \cite{opsahl-ong-etal-2024-optimizing} extends this bootstrapping paradigm by jointly optimizing both proposed instruction text and demonstration selection using Bayesian search, thereby expanding the prompt optimization space.}

For fine-tuning, \textbf{BootstrapFinetune} \cite{soylu-etal-2024-bettertogether, khattab2024dspy} iteratively solves training examples from \(\mathcal{D}_{\mathrm{train}}\), generating solution traces which are curated and used to further fine-tune the model, e.g., via LoRA adapters \cite{hu2022lora}, enabling a self-improving loop.

Following the BetterTogether strategy, {we study hybrid pipelines that alternate between bootstrap-based prompt optimizers} and parameter-efficient fine-tuning, yielding mutual improvements in prompt quality and model parameters.

\begin{algorithm}[t]
\caption{{\textbf{ClusterFewshot}: Semantic-Aware Bootstrap Demonstration Selection}}
\label{alg:clusterfewshot}
\begin{algorithmic}

\Require
Training data: $\mathcal{D}_{\mathrm{train}}, \mathcal{D}_{\mathrm{val}}$ \\
Program: $\mathcal{P}_\theta$ \\
Candidate embedders: $\mathcal{M}=\{M_j\}_{j=1}^J$ \\
Cluster range: $K \in [K_{\min},K_{\max}]$ \\
Demo budget: $k$ \\
Validation samples per cluster: $m$

\Ensure
Selected demonstrations $\mathcal{F}^*$

\State \textbf{Bootstrapping Training Set}
\State $\tilde{\mathcal{D}}_{\mathrm{train}} \gets \mathrm{Bootstrap}(\mathcal{P}_\theta,\mathcal{D}_{\mathrm{train}})$

\State \textbf{Embedding and clustering selection}
\For{$M_j \in \mathcal{M}$}
    \State $\mathbf{e}_i^{(j)} \gets M_j(x_i)$ for all $x_i \in \tilde{\mathcal{D}}_{\mathrm{train}}$
    \For{$K \in [K_{\min},K_{\max}]$}
        \State $\hat{y}^{(j,K)} \gets \mathrm{KMeans}_K(\{\mathbf{e}_i^{(j)}\})$
        \State $S_{j,K} \gets \mathrm{Silhouette}(\{\mathbf{e}_i^{(j)}\},\hat{y}^{(j,K)})$
    \EndFor
\EndFor
\State $(j^*,K^*) \gets \arg\max_{j,K} S_{j,K}$

\State \textbf{Cluster training examples}
\Statex \parbox[t]{\linewidth}{
$\{\mathcal{C}_k^{\mathrm{train}}\}_{k=1}^{K^*}
\gets
\mathrm{KMeans}_{K^*}\!\big(M_{j^*}(\tilde{\mathcal{D}}_{\mathrm{train}})\big)$
}

\State \textbf{Construct one-shot evaluation subset}
\Statex \parbox[t]{\linewidth}{
$\{\mathcal{C}_k^{\mathrm{val}}\}_{k=1}^{K^*}
\gets
\mathrm{KMeans}_{K^*}\!\big(M_{j^*}(\mathcal{D}_{\mathrm{val}})\big)$
}

\Statex $\boldsymbol{\mu}_c \gets \mathrm{centroid}\!\left(\mathcal{C}_c^{\mathrm{val}}\right),
\quad \forall c \in \{1,\dots,K^*\}$

\State $\mathcal{C}_{\mathrm{val}} \gets
\bigcup_{c=1}^{K^*}
\arg\min_{x \in \mathcal{C}_c^{\mathrm{val}}}^{(m)}
\| M_{j^*}(x) - \boldsymbol{\mu}_c \|$

\Statex \Comment{Select $m$ nearest validation examples to each cluster centroid in embedding space}

\State \textbf{One-shot scoring}
\For{$x_i \in \tilde{\mathcal{D}}_{\mathrm{train}}$}
    \State $s_i \gets \frac{1}{|\mathcal{C}_{\mathrm{val}}|}
    \sum_{(x,y)\in\mathcal{C}_{\mathrm{val}}}
    \mathrm{Eval}(\mathcal{P}_\theta(\{x_i\},x),y)$
\EndFor

\State \textbf{Candidate construction}
\State $\mathcal{F}_{\textsc{Top-}k} \gets \mathrm{Top}_k(\{x_i\}, s_i)$
\State $\mathcal{F}_{\textsc{Cluster}} \gets \bigcup_{c=1}^{K^*}
\arg\max_{x_i \in \mathcal{C}_c^{\mathrm{train}}} s_i$

\State \textbf{Final selection}
\State $\mathcal{F}_{\mathrm{cand}} \gets \{\mathcal{F}_{\textsc{Top-}k},\mathcal{F}_{\textsc{Cluster}}\}$
\State $\mathcal{F}^* \gets \arg\max\limits_{\mathcal{F}\in\mathcal{F}_{\mathrm{cand}}}
\mathrm{Score}(\mathcal{P}_\theta[\mathcal{F}],\mathcal{D}_{\mathrm{val}})$

\State \Return $\mathcal{F}^*$

\end{algorithmic}
\end{algorithm}

\section{Method}
\label{Method}
We propose \textbf{ClusterFewshot}, our semantically informed bootstrap‐selection procedure.  ClusterFewshot replaces random sampling or metric‐based ranking with a two‐stage process: (i) example embedding‐based clustering, and (ii) evaluation‐driven selection across sampling strategies.

Similar to other bootstrap-based approaches \autoref{Background},
our goal is to select a small bootstrapped demonstration set from
$\mathcal{D}_{\mathrm{train}}$ of size $k$ that maximizes the validation performance of our LLM workflow $\mathcal{P}_\theta$.

While clustering-based sampling has been explored in prior work, our focus is on how it can be integrated with utility-based signals to improve the efficiency and stability of LLM optimization pipelines by organizing demonstrations according to the latent semantic structure of the task. Using this design of complementary signals encourages coverage of the latent task structure, while utility-aware scoring prioritizes demonstrations that are empirically useful for the target model and task.

\subsection{Bootstrapping Training Set}
\label{Bootstrapped Training Set}

{We construct a bootstrapped training set $\tilde{\mathcal{D}}_{\mathrm{train}}$ by executing the LM program $\mathcal{P}_\theta$ on the full training set $\mathcal{D}_{\mathrm{train}}$
and retaining successful trace-derived demonstrations.
Unlike prior approaches that repeatedly bootstrap {from} randomly sampled subsets during search \cite{khattab2024dspy, opsahl-ong-etal-2024-optimizing}, ClusterFewshot performs this step once and subsequently operates only on $\tilde{\mathcal{D}}_{\mathrm{train}}$.}

\subsection{Semantic Embedding and Clustering}
\label{Semantic Embedding and Clustering}
To capture the semantic diversity of {$\tilde{\mathcal{D}}_{\mathrm{train}}$}, we embed each input $x_i$ using a candidate model $M_j$, either a pretrained sentence encoder or the task-tuned LLM, yielding an embedding $\mathbf{e}_i^{(j)} \in \mathbb{R}^d$, where $d$ is the embedding dimension.
We then apply K‐means clustering \cite{lloyd1982least} to $\left\{ \mathbf{e}_i^{(j)} \right\}$, to maximize the Silhouette score \cite{rousseeuw1987silhouettes}:
\[
    \left\{ \mathbf{e}_i^{(j)} = M_j(x_i) \right\}_{i=1}^N \ \text{for } M_j \in \mathcal{M},\; j = 1, \dots, J
\]
The selection procedure uses a grid‐search calibration phase, with additional analysis in \autoref{sec:cluster_embedding_analysis}:
\begin{align}
    \hat{y}^{(j,k)} &= \mathrm{KMeans}_k\left( \left\{ \mathbf{e}_i^{(j)} \right\}_{i=1}^N \right) \label{eq:kmeans} \\
    S_{j,k} &= \mathrm{Silhouette}\left( \left\{ \mathbf{e}_i^{(j)} \right\},\; \hat{y}^{(j,k)} \right) \label{eq:silhouette} \\
    (j^*, k^*) &= \arg\max_{\substack{j \in [1, J] \\ k \in [K_{\min}, K_{\max}]}} S_{j,k} \label{eq:opt_selection}
\end{align}
\noindent
This grid search also selects one of the following sentence-transformers:
\begin{itemize}

    \item \textbf{all-mpnet-base-v2} \cite{reimers2019sentencebertsentenceembeddingsusing} –
    {Transformer-based sentence embedding model, built on MPNet, that produces 768-dimensional vector representations of sentences and short paragraphs, and is widely used for similarity and semantic search tasks.}
    
    \item \textbf{gtr-t5-base} \cite{raffel2020exploring} –
    {a T5-based sentence embedding model designed for semantic search and retrieval, producing high-quality, cross-lingual dense vector representations of sentences and paragraphs.}

    \item \textbf{bge-large-en-v1.5} \cite{bge-large-en-v1.5} -
    {a bidirectional-encoder model optimized for English dense retrieval via contrastive learning on web-scale corpora.}

    \item \textbf{Qwen3-Embedding-0.6B} \cite{qwen3embedding} -
    {an embedding model from the Qwen3 series, producing fixed-dimensional representations via fine-tuned language model layers for retrieval and reranking tasks.}

\end{itemize}

\noindent
Only for Iris, given the well-structured numerical nature of the task input, we use the original feature vectors, which comprise features such as {sepal length and width}.

\subsection{One‐Shot Candidate Scoring}
To estimate the utility of each candidate example {\(x_i \in \tilde{\mathcal{D}}_{\mathrm{train}}\)} individually, we define a small evaluation subset \(\mathcal{C}_{\mathrm{val}} \subset \mathcal{D}_{\mathrm{val}}\). This subset is constructed by first clustering \(\mathcal{D}_{\mathrm{val}}\) in the semantic embedding space, then selecting {the \(m\)} examples closest to each cluster centroid, yielding a diverse and representative evaluation set. {For consistency, we fix the number of validation samples per cluster to \(m=3\).}
{We then compute a one-shot score \(s_i\) for each candidate \(x_i\)} by measuring its effectiveness as the sole demonstration when prompting the LM on \(\mathcal{C}_{\mathrm{val}}\). Specifically, we define:
\begin{align}
s_i = \frac{1}{|\mathcal{C}_{\mathrm{val}}|} \sum_{(x, y) \in \mathcal{C}_{\mathrm{val}}} \mathrm{Eval}\left(\mathcal{P}_\theta(\{x_i\},\,x),\,y\right)
\end{align}
where:
  \(\mathcal{P}_\theta(\{x_i\},\,x)\) denotes the model’s prediction on input \(x\) when prompted with \(x_i\) as the sole in-context demonstration.
  \(\mathrm{Eval}(\cdot, y)\) is a task-specific evaluation metric comparing the model’s output to the ground-truth label \(y\).
This ``one‐shot evaluation'' rapidly ranks examples by their individual contribution to the LM performance on a diverse set of questions,  {replacing budget-limited subset search \cite{opsahl-ong-etal-2024-multistage, soylu-etal-2024-bettertogether} with linear per-candidate evaluation under fixed validation cost to enable a tractable and exhaustive scoring of all training candidates from \(\tilde{\mathcal{D}}_{\mathrm{train}}\).}

\subsection{Sampling Strategies}
We employ a sampling framework that combines global and cluster-based selection, adapting to the task's nature. This design balances high-performing examples with semantic diversity in the resulting few-shot subset.

For Iris, a classification task, we prioritize class-level cluster coverage by selecting the top-performing demonstration from each class.

Accordingly, we adopt the following strategies: 
  (i) \textbf{Global Top‐$k$:} select the $k$ examples with highest $s_i$ across all clusters.
  (ii) \textbf{Cluster Representatives:} select, for each cluster $\mathcal{C}_k$, the example with highest $s_i$ within that cluster.
Each strategy $S$ yields a candidate set $\mathcal{F}_S$ of size $\le k$.

\subsection{Final Selection and Compilation}
Finally, we evaluate each candidate $\mathcal{F}_S$ on the full validation set and choose
\begin{align}
\mathcal{F}^* = \arg\max_{S}
\mathrm{Score}\bigl(\mathcal{P}_\theta[\mathcal{F}_S],\,\mathcal{D}_{\mathrm{val}}\bigr)
\end{align}
The optimized program $\mathcal{P}_\theta[\mathcal{F}^*]$ is then ready for inference. {When used within BetterTogether’s interleaved optimization approach \cite{soylu-etal-2024-bettertogether}}, this phase may be followed by an additional fine-tuning phase.

\begin{table*}[h]
\centering
\small
\renewcommand{\arraystretch}{1.1}
\setlength{\tabcolsep}{5.8pt}
\begin{tabular}{llc ccc ccc}
\toprule
\textbf{Strategy} & \textbf{Prompt Optimizer} & \textbf{Opt.} &
\multicolumn{3}{c}{\textbf{Qwen2.5-7B-Instruct}} &
\multicolumn{3}{c}{\textbf{Llama-3.2-3B-Instruct}} \\
& & & GSM8K & HotPotQA & Iris & GSM8K & HotPotQA & Iris \\
\midrule

\multirow{3}{*}{P}
& BFRS           & D   & 85.59 & 45.98 & 78.00 & 78.24 & \textbf{36.95} & 69.33 \\
& MIPROv2        & I+D & 84.82 & 44.93 & 79.33 & 75.97 & 35.25 & 70.66 \\
& ClusterFewshot & D   & \textbf{87.78} & \textbf{46.29} & \textbf{83.33} & \textbf{79.54} & 36.84 & \textbf{71.33} \\
\midrule

\multirow{3}{*}{P $\rightarrow$ P}
& BFRS           & D   & 86.11 & 43.71 & 82.66 & 78.27 & 40.69 & 68.00 \\
& MIPROv2        & I+D & 82.80 & 45.18 & 78.66 & 77.39 & 34.82 & 62.66 \\
& ClusterFewshot & D   & \textbf{87.86} & \textbf{45.33} & \textbf{86.66} & \textbf{81.05} & \textbf{40.93} & \textbf{75.33} \\
\midrule

\multirow{3}{*}{W $\rightarrow$ P}
& BFRS           & D   & 87.17 & 42.47 & 80.00 & 78.32 & 35.97 & 69.33 \\
& MIPROv2        & I+D & 83.06 & \textbf{44.62} & \textbf{84.00} & 75.32 & 31.75 & 60.66 \\
& ClusterFewshot & D   & \textbf{87.88} & 42.84 & 80.66 & \textbf{80.17} & \textbf{37.35} & \textbf{72.66} \\
\midrule

\multirow{3}{*}{P $\rightarrow$ W}
& BFRS           & D   & 85.48 & 44.13 & 84.00 & 76.50 & \textbf{38.29} & 68.66 \\
& MIPROv2        & I+D & 79.56 & 43.85 & 83.33 & 75.14 & 29.78 & 68.00 \\
& ClusterFewshot & D   & \textbf{86.16} & \textbf{44.22} & \textbf{85.33} & \textbf{77.66} & 36.82 & \textbf{71.33} \\
\midrule

\multirow{3}{*}{P $\rightarrow$ W $\rightarrow$ P}
& BFRS           & D   & 82.09 & 45.13 & \textbf{84.66} & 78.42 & 36.26 & 66.66 \\
& MIPROv2        & I+D & 77.62 & 43.35 & 84.00 & 73.75 & 31.40 & 68.00 \\
& ClusterFewshot & D   & \textbf{86.01} & \textbf{45.48} & 82.00 & \textbf{79.11} & \textbf{38.51} & \textbf{75.33} \\
\bottomrule
\end{tabular}
\caption{
Evaluation of BetterTogether strategies 
using Qwen2.5-7B-Instruct and Llama-3.2-3B-Instruct.
\textbf{Opt.} denotes optimized prompt components:
\textbf{D} = demonstrations only (BFRS, ClusterFewshot),
\textbf{I+D} = joint instruction and demonstration optimization (MIPROv2).
\textbf{P} denotes prompt optimization and \textbf{W} denotes parameter-efficient
weight tuning.
Reported values are averages over three independent runs.
}
\vspace{5px}
\label{tab:bt_all_models_qwen_llama}
\end{table*}

\begin{table*}[h]
\centering
\small
\renewcommand{\arraystretch}{1.1}
\setlength{\tabcolsep}{6.2pt}
\begin{tabular}{llcccccc}
\toprule
\textbf{Strategy} & \textbf{Prompt Optimizer} &
\multicolumn{3}{c}{\textbf{Qwen2.5-7B-Instruct}} &
\multicolumn{3}{c}{\textbf{Llama-3.2-3B-Instruct}} \\
& & GSM8K & HotPotQA & Iris & GSM8K & HotPotQA & Iris \\
\midrule

\multirow{3}{*}{P}
& BFRS           & 23.58 & 42.88 & 1.51 & 12.39 & 28.22 & 0.71 \\
& MIPROv2        & 19.76 & 59.11 & 2.84 & 13.98 & 104.63 & 1.63 \\
& ClusterFewshot & \textbf{18.39} & \textbf{25.48} & \textbf{0.47} & \textbf{9.69} & \textbf{19.61} & \textbf{0.24} \\
\midrule

\multirow{3}{*}{P $\rightarrow$ P}
& BFRS           & 39.92 & 71.63 & 2.50 & 20.00 & 50.83 & 1.28 \\
& MIPROv2        & 37.52 & 149.36 & 8.28 & 30.71 & 172.21 & 4.07 \\
& ClusterFewshot & \textbf{22.62} & \textbf{37.81} & \textbf{0.81} & \textbf{15.24} & \textbf{23.37} & \textbf{0.44} \\
\midrule

\multirow{3}{*}{W $\rightarrow$ P}
& BFRS           & 44.08 & 71.79 & 5.29 & 23.89 & 39.38 & 2.09 \\
& MIPROv2        & 42.16 & 80.94 & 5.04 & 28.40 & 71.24 & 3.05 \\
& ClusterFewshot & \textbf{41.78} & \textbf{57.15} & \textbf{4.01} & \textbf{21.13} & \textbf{32.12} & \textbf{1.65} \\
\midrule

\multirow{3}{*}{P $\rightarrow$ W}
& BFRS           & 43.04 & 80.45 & 5.21 & 26.90 & 46.39 & 2.28 \\
& MIPROv2        & 39.81 & 91.37 & 5.07 & 27.38 & 89.13 & 3.20 \\
& ClusterFewshot & \textbf{36.65} & \textbf{61.61} & \textbf{3.70} & \textbf{22.98} & \textbf{35.80} & \textbf{1.98} \\
\midrule

\multirow{3}{*}{P $\rightarrow$ W $\rightarrow$ P}
& BFRS           & 77.70 & 111.36 & 6.46 & 32.30 & 64.10 & 2.96 \\
& MIPROv2        & 65.38 & 187.88 & 9.99 & 40.85 & 165.19 & 5.44 \\
& ClusterFewshot & \textbf{46.61} & \textbf{75.97} & \textbf{5.14} & \textbf{28.80} & \textbf{44.22} & \textbf{2.14} \\
\bottomrule
\end{tabular}

\caption{
Runtime comparison of BetterTogether strategies across GSM8K, HotPotQA, and Iris benchmarks using Qwen2.5-7B-Instruct and Llama-3.2-3B-Instruct.
{Reported values are average end-to-end runtimes (in minutes) over three independent runs, and they reflect the full optimization pipeline, including prompt compilation, fine-tuning where applicable, and evaluation.}
}
\label{tab:runtime_comparison_qwen_llama}
\end{table*}

\section{Experiments}
\label{sec:experiments}
{In this section we evaluate ClusterFewshot as an adaptive and efficient optimization method.}

\subsection{BetterTogether Evaluation}
\label{sec:BetterTogether Evaluation}
We evaluate our approach on the benchmarks introduced in the original BetterTogether study, covering both individual optimization phases and interleaved pipelines with fine-tuning:
\begin{itemize}
  \item \textbf{GSM8K} \cite{cobbe2021gsm8k}: Grade‑school math word problems requiring multi‑step arithmetic reasoning. This task is implemented via an LM program with a single chain-of-thought prompting module. The program uses 1,000 training and all available 1,319 test examples, drawn from the original training and test sets. For prompt optimization, 100 training and 250 validation examples are sub-sampled, without overlap. Accuracy is measured by extracting the last number from the model's first-line response and comparing it to the ground truth.
  \item \textbf{HotPotQA} \cite{yang2018hotpotqa}: Using a large corpus of Wikipedia abstracts, {identify two passages that together contain the factual evidence needed to answer a multi-hop question}. This task is implemented via an LM program with three chain-of-thought modules, arranged in a multi-hop pipeline. The program uses 1,000 training and 1,500 test examples, sampled from the original training and validation sets. For prompt optimization, 100 training and 250 validation examples are sub-sampled, without overlap. Accuracy is evaluated using exact match.
  \item \textbf{Iris} \cite{fisher1936use}: a classification task {over} iris flowers, based on the given sepal/petal dimension features. This task is implemented via an LM program with a single chain-of-thought module. It uses 50 training and 50 test examples, sampled from the original Iris dataset. For prompt optimization, 15 training and 35 validation examples are sub-sampled, without overlap. Accuracy is evaluated using exact match.
\end{itemize}

For \textbf{BFRS}, we explore six candidate LM program configurations per run - covering zero-shot (Vanilla), labeled-only, and several bootstrapped few-shot variants. This setup aligns with that of the original BetterTogether paper \cite{soylu-etal-2024-bettertogether}.

\begin{figure}[h]
    \centering
    \includegraphics[width=0.91\linewidth]{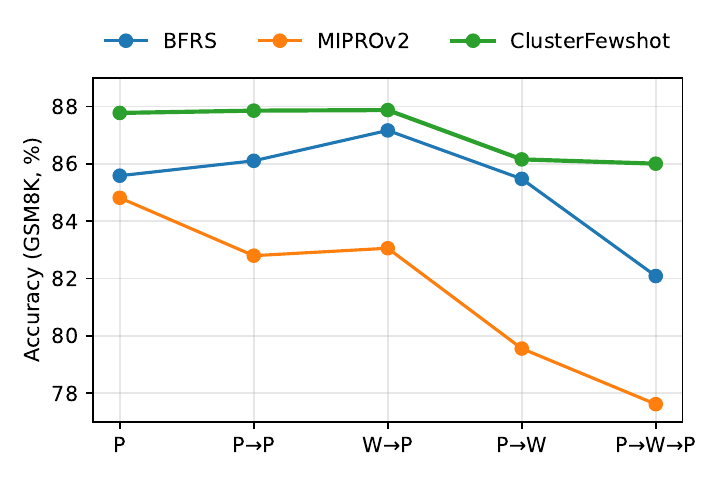}
    \caption{
    Accuracy across BetterTogether strategies
    on GSM8K using Qwen2.5-7B-Instruct.
    }
    \label{fig:bt_gsm8k_qwen_trends_acc}
\end{figure}

\begin{figure}[h]
    \centering
    \includegraphics[width=0.91\linewidth]{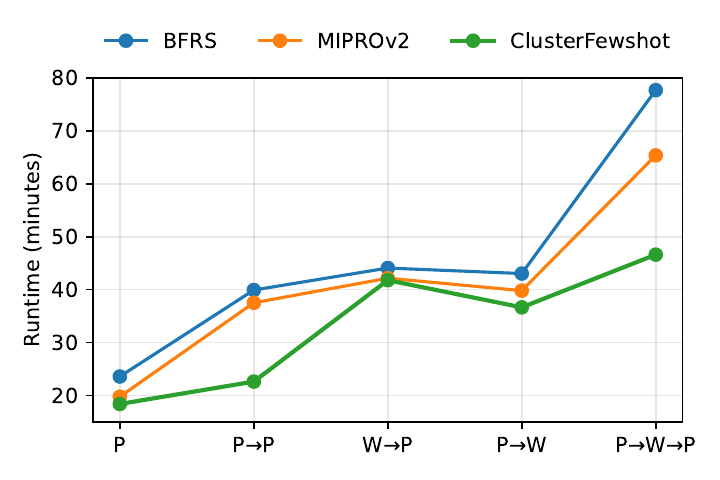}
    \caption{
    End-to-end runtime across BetterTogether strategies
    on GSM8K using Qwen2.5-7B-Instruct.
    }
    \label{fig:bt_gsm8k_qwen_trends_runtime}
\end{figure}

\begin{figure}[h]
    \centering
    \includegraphics[width=0.87\linewidth]{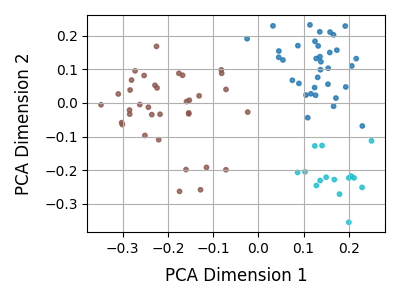}
    \caption{
    PCA projection of \textbf{training} example embeddings on GSM8K
    using the \textbf{gtr-t5-base} encoder.
    We cluster $85$ training examples into $k=3$ semantic groups using k-means.
    }
    \label{fig:clusters_gsm8k_train}
\end{figure}

\begin{figure}[h]
    \centering
    \includegraphics[width=0.87\linewidth]{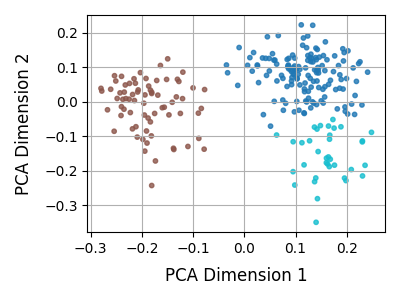}
    \caption{
    PCA projection of $250$ \textbf{validation} example embeddings on GSM8K using the \textbf{gtr-t5-base} encoder.
    }
    \label{fig:clusters_gsm8k_val}
\end{figure}

For \textbf{MIPROv2}, we use the standard \emph{medium} configuration, which proposes 12 instruction candidates and 12 bootstrapped few-shot demonstration sets per optimization run \cite{opsahl-ong-etal-2024-optimizing}.

Leveraging DSPy’s compatibility with SGLang, a high-performance serving backend for structured and compositional LLM generation \cite{zheng2024sglangefficientexecutionstructured}, we conduct our experiments with Qwen-2.5-7B \cite{qwen2.5} and Llama-3.2-3B \cite{llama3.2}.

Table~\ref{tab:bt_all_models_qwen_llama} shows a clear advantage for ClusterFewshot. It outperforms both BFRS and MIPROv2 across all three benchmarks in the majority of standalone and hybrid BetterTogether configurations. On \textbf{GSM8K} and \textbf{Iris}, ClusterFewshot delivers consistent improvements across most strategies, indicating enhanced suitability for arithmetic reasoning and structured classification tasks. It also achieves competitive or superior performance on \textbf{HotPotQA}, suggesting improved robustness in complex multi-hop reasoning settings.
{This trend is illustrated in \autoref{fig:bt_gsm8k_qwen_trends_acc}, where ClusterFewshot consistently achieves the strongest GSM8K performance with Qwen2.5-7B-Instruct across strategies.}

\begin{table*}[!htbp]
\centering
\small
\setlength{\tabcolsep}{4.2pt}
\renewcommand{\arraystretch}{1.08}
\begin{tabular}{lccc ccc ccc}
\toprule
\multirow{2}{*}{\textbf{Optimizer}} &
\multicolumn{3}{c}{\textbf{GSM8K}} &
\multicolumn{3}{c}{\textbf{HotPotQA}} &
\multicolumn{3}{c}{\textbf{Iris}} \\
\cmidrule(lr){2-4}
\cmidrule(lr){5-7}
\cmidrule(lr){8-10}
&
\textbf{Tokens} & \textbf{Compile} & \textbf{Score} &
\textbf{Tokens} & \textbf{Compile} & \textbf{Score} &
\textbf{Tokens} & \textbf{Compile} & \textbf{Score} \\
&
& \textbf{(min)} & \textbf{\(\Delta\)} &
& \textbf{(min)} & \textbf{\(\Delta\)} &
& \textbf{(min)} & \textbf{\(\Delta\)} \\
\midrule
BFRS
& 2.11M & 15.6 & +1.0pp
& 8.75M & 31.3 & +13.9pp
& 259K & 1.3 & +36pp \\
MIPROv2
& \textbf{1.91M} & 16.5 & \textbf{+7.7pp}
& 7.19M & 25.5 & +13.9pp
& 910K & 5.2 & \textbf{+46pp} \\
ClusterFewshot
& 2.21M & \textbf{14.7} & +7.5pp
& \textbf{6.31M} & \textbf{13.3} & \textbf{+19.1pp}
& \textbf{226K} & \textbf{0.8} & \textbf{+46pp} \\
\bottomrule
\end{tabular}
\caption{{
Token usage, compilation time, and score improvements across prompt optimizers and tasks using Qwen2.5-7B-Instruct.
}}
\label{tab:token_runtime_efficiency_qwen7b}
\end{table*}

Extended results in Appendix~\ref{sec:extended_results} show that ClusterFewshot exhibits lower variance than both BFRS and MIPROv2 across most strategies, indicating improved stability across independent runs.

Runtime comparisons are reported in Table~\ref{tab:runtime_comparison_qwen_llama}, and illustrated in \autoref{fig:bt_gsm8k_qwen_trends_runtime}. ClusterFewshot achieves substantially lower runtime than both random and Bayesian search-based prompt optimization approaches across all prompt-only and hybrid configurations, reducing latency on \textbf{HotPotQA} by over $\mathbf{40\%}$ in the standalone setting, while maintaining strong performance in multi-stage workflows. These results highlight ClusterFewshot’s favorable balance between accuracy, stability, and computational efficiency across open-source model families, including smaller and medium-scale models.

To further contextualize these results, Figures~\ref{fig:clusters_gsm8k_train} and~\ref{fig:clusters_gsm8k_val} visualize the semantic structure induced by the embedding space on \textbf{GSM8K}. The observed consistency across different splits supports the use of embedding-based clustering for both identifying representative regions for constructing diverse few-shot candidate pools from $\tilde{\mathcal{D}}_{\mathrm{train}}$, and sampling informative subsets from $\mathcal{D}_{\mathrm{val}}$ for downstream one-shot utility estimation.


\subsection{Compilation Efficiency}
\label{sec:compilation_efficiency}

{While Section~\ref{sec:BetterTogether Evaluation} reports end-to-end execution time, which includes both the optimization phase and test evaluation, we further isolate the compilation stage efficiency of each optimizer. We measure this cost using compilation wall-clock  time and token usage as practical indicators.}

Table~\ref{tab:token_runtime_efficiency_qwen7b} compares overall token usage, compilation time, and score improvements across GSM8K, HotPotQA, and Iris using \textbf{Qwen2.5-7B-Instruct}. Across all tasks, ClusterFewshot exhibits a favorable optimization-efficiency profile, achieving the lowest compile time on Iris and HotPotQA while remaining competitive on GSM8K. The token-usage results further indicate that ClusterFewshot obtains these runtime gains with a generally lower token cost. Among the compared optimizers, ClusterFewshot uses the fewest tokens on HotPotQA and Iris, while maintaining a comparable token usage on GSM8K.

{Appendix~\ref{sec:cluster_embedding_analysis} further analyzes the efficiency of ClusterFewshot's underlying components, focusing on its embedding and clustering stages, showing how these components balance semantic separation quality with computational efficiency.}

{
Overall, these results support that revealing the semantic structure of a task provides an effective control signal for demonstration selection.
}

\begin{table*}[t]
\centering
\footnotesize
\begin{tabular}{lcccccc}
\toprule
\textbf{Optimizer}
& \multicolumn{3}{c}{\textbf{Qwen2.5-7B-Instruct}}
& \multicolumn{3}{c}{\textbf{Qwen2.5-14B-Instruct}} \\
\cmidrule(lr){2-4}
\cmidrule(lr){5-7}
& \textbf{Accuracy} & \textbf{acc@$\leq$2} & \textbf{Compile}
& \textbf{Accuracy} & \textbf{acc@$\leq$2} & \textbf{Compile} \\
& & & \textbf{(min.)} & & & \textbf{(min.)} \\
\midrule
Baseline
& $24.53$ & $23.38$ & -- 
& $47.80$ & $44.35$ & -- \\
ClusterFewshot
& $\mathbf{45.87}\pm2.89$ & $\mathbf{45.79}$ & $\mathbf{14.1}$
& $\mathbf{54.65}\pm1.93$ & $\mathbf{52.63}$ & $\mathbf{55.1}$ \\
BFRS
& $44.95\pm1.48$ & $44.70$ & $38.9$
& $53.76\pm1.22$ & $49.40$ & $75.1$ \\
MIPROv2
& $44.80\pm4.04$ & $43.89$ & $35.6$
& $52.22\pm0.24$ & $52.57$ & $69.6$ \\
\bottomrule
\end{tabular}
\caption{
{Agentic HotPotQA results using \texttt{dspy.ReAct} with ColBERTv2 retrieval.
Accuracy is averaged over three seeds. acc@$\leq$2 denotes
accuracy on trajectories that terminate within two ReAct steps. Compile time is measured as mean runtime in minutes.}
}
\label{tab:agentic_react_hotpotqa}
\end{table*}

\subsection{ReAct Agentic Setting}
\label{sec:agentic_setting}

To assess whether cluster-guided demonstration selection generalizes beyond
standard Chain-of-Thought prompting, we further evaluate tool-augmented agentic setting: we instantiate \textbf{\texttt{ReAct}}~\cite{yao2023reactsynergizingreasoningacting} on
HotPotQA multi-hop question answering with \textbf{ColBERTv2} Wikipedia retrieval~\cite{santhanam2022colbertv2effectiveefficientretrieval}, retrieving three passages per query. In this setting, the model interleaves reasoning and retrieval across multiple steps and must explicitly terminate with a \texttt{Finish[]} action for its answer to be evaluated. Runs that reach the maximum step limit without emitting \texttt{Finish[]} are marked
incorrect.
Each trajectory is allowed up to $20$ ReAct steps, and each
optimizer compiles up to four demonstrations. Results are averaged over three seeds on 1,500 test examples, using Qwen2.5 
at both 7B and 14B scales.

Table~\ref{tab:agentic_react_hotpotqa} reports overall accuracy, early-termination accuracy, and mean compilation time. The main observation is that ClusterFewshot consistently reduces compilation time across model scales while preserving, and even improving, mean accuracy. The reduction is most pronounced at 7B, where ClusterFewshot lowers compilation time by $\mathbf{63.8\%}$ relative to BFRS and $\mathbf{60.4\%}$ relative to MIPROv2. ClusterFewshot also obtains the highest mean accuracy and early-termination accuracy at both scales, exhibiting efficient answer
completion early in the ReAct trajectory.

This efficiency highlights the practical applicability of ClusterFewshot in
real-world agentic deployments, where in-context demonstrations may need to be
continuously re-compiled across models, tools, and retrieval corpora as task
distributions evolve. In such dynamic settings, ClusterFewshot provides an
efficient way to obtain high-quality, representative demonstrations that guide not only final-answer reasoning but also intermediate tool-use and reasoning steps.

\subsection{Individual Sampling Strategies and Retrieval Performance}
\label{sec:sampling_retrieval_analysis}

To provide a more comprehensive analysis of ClusterFewshot's design components, we analyze different cluster sampling strategies and evaluate our approach against retrieval-based methods.

Table~\ref{tab:individual_sampling} analyzes individual cluster sampling strategies and situates them relative to full prompt optimizers. These variants consist of global Top-$k$ selection, cluster representatives, centroid-based sampling, and cluster-level random selection. While each strategy captures either semantic diversity or utility-based signals, none consistently matches the hybrid approach across tasks. The hybrid variant, which selects between Top-$k$ and representative sampling based on one-shot validation performance, achieves the strongest results, supporting the core design of ClusterFewshot.

We further compare ClusterFewshot to \textbf{RetrievalFewshot (RFS)}, a query-adaptive baseline that selects demonstrations at inference time from the same bootstrapped candidate pool using semantic retrieval over the embedding space. Table~\ref{tab:retrieval_comparison} reports results for kNN retrieval, which selects the nearest examples to the given query, and Maximal Marginal Relevance (MMR), which balances relevance and diversity in the resulting few-shot through a trade-off parameter $\lambda$ (higher $\lambda$ favors similarity, lower $\lambda$ encourages diversity). Unlike these query-adaptive approaches, ClusterFewshot compiles a fixed few-shot context during optimization and reuses it for all queries. Despite using semantic similarity directly at query time, these retrieval-based strategies consistently underperform ClusterFewshot.

These findings indicate that compiling a fixed few-shot context that {combines semantic and utility-based signals} yields the strongest performance {among these variants}.
Extended results, including statistical variability across multiple runs of Tables~\ref{tab:individual_sampling} and \ref{tab:retrieval_comparison}, are provided in \autoref{sec:rebuttal_ablations}, demonstrating the consistency of our findings. We further include a zero-shot evaluation in \autoref{sec:rebuttal_ablations}, where ClusterFewshot provide consistent gains across tasks.

\begin{table}[t]
\centering
\footnotesize
\begin{tabular}{lccc}
\toprule
\textbf{Method} & \textbf{GSM8K} & \textbf{HotPotQA} & \textbf{Iris} \\
\midrule
CFS - Global Top-$k$  & $86.11$ & $42.20$ & $77.47$ \\
CFS - Representatives & $85.98$ & $44.97$ & $86.27$ \\
CFS - Cluster Random  & $85.24$ & $41.29$ & $76.00$ \\
CFS - Centroids       & $86.39$ & $43.22$ & $79.60$ \\
\midrule
BFRS                              & $84.68$ & $44.48$ & $80.40$ \\
MIPROv2                           & $84.17$ & $44.21$ & $84.40$ \\
\midrule
\textbf{CFS -- Hybrid} & $\mathbf{88.06}$ & $\mathbf{46.41}$ & $\mathbf{86.27}$ \\
\bottomrule
\end{tabular}
\caption{Performance of individual ClusterFewshot (CFS) sampling strategies variants compared with full optimizers.}
\label{tab:individual_sampling}
\end{table}

\begin{table}[t]
\centering
\footnotesize
\begin{tabular}{lccc}
\toprule
\textbf{Method} & \textbf{GSM8K} & \textbf{HotPotQA} & \textbf{Iris} \\
\midrule
RFS-kNN                   & $84.70$ & $42.71$ & $76.67$ \\
RFS-MMR ($\lambda{=}0.2$) & $84.19$ & $43.38$ & $72.00$ \\
RFS-MMR ($\lambda{=}0.5$) & $83.18$ & $44.20$ & $76.67$ \\
RFS-MMR ($\lambda{=}0.8$) & $83.13$ & $43.20$ & $77.33$ \\
\midrule
\textbf{ClusterFewshot}   & $\mathbf{87.78}$ & $\mathbf{46.29}$ & $\mathbf{83.33}$ \\
\bottomrule
\end{tabular}
\caption{Comparison with retrieval-based few-shot baselines that select demonstrations at inference time using semantic similarity.}
\label{tab:retrieval_comparison}
\end{table}

\section{Conclusions}
We presented \textbf{ClusterFewshot}, a semantically guided bootstrap-selection method that integrates embedding-based clustering with one-shot evaluation and multi-strategy sampling.

Our extended evaluation shows that ClusterFewshot improves the efficiency of reasoning-augmented demonstration selection by substantially reducing optimization {costs} across diverse models and tasks, while consistently preserving strong downstream accuracy.

These improvements extend beyond standalone settings to modern multi-stage LLM optimization workflows such as BetterTogether {and agentic ReAct, as shown in Tables~\ref{tab:bt_all_models_qwen_llama},~\ref{tab:runtime_comparison_qwen_llama} and~\ref{tab:agentic_react_hotpotqa}}, with further analysis in Section~\ref{sec:extended_results}.

{ClusterFewshot shows robustness across embedding choices and tasks, exhibiting stable and efficient semantic structuring.}

{Looking forward, }its modular design and reliance on off-the-shelf embedding encoders make it readily adaptable to new tasks, model families, and dynamic data regimes.

\section*{Limitations}
While \textbf{ClusterFewshot} yields consistent improvements across multiple tasks and strategies, several limitations warrant consideration.

First, the method assumes that off-the-shelf embeddings capture task-relevant semantics. In domains where this alignment is weak, induced clusters may fail to meaningfully partition training or validation data, reducing demonstration quality.

Second, estimating the number of clusters via Silhouette maximization introduces an additional hyperparameter to tune. On large datasets, repeated K-means runs over multiple embeddings and candidate \( K \) values may partially offset the method’s overall efficiency gains.

Third, one-shot evaluation depends on a held-out subset \(\mathcal{C}_{\mathrm{val}}\), which may be small or unrepresentative. In such cases, selected demonstrations may not generalize despite local performance gains.

Finally, although we use \textbf{BootstrapFewShotRS (BFRS)} and \textbf{MIPROv2} as our main baselines, future work may extend this comparison to additional semantically driven selection techniques and hybrid optimization methods. Further directions might also include adaptive clustering that evolves with streaming examples, integration with active-learning loops to minimize human annotation effort, additional tool calling in agentic environments, and theoretical analysis of cluster coherence under fine-tuning, which we did not cover in this work.

\smallskip
\smallskip
\section*{Ethics Statement}
This work proposes \textbf{ClusterFewshot}, a method for selecting high-quality few-shot demonstrations to enhance the performance and generalization of language models. The method is model-agnostic and leverages semantic clustering and one-shot evaluation to improve prompt optimization efficiency.

We acknowledge that techniques improving the effectiveness and reliability of language models can be dual-use. While our approach is intended to support responsible and beneficial applications, such as improved reasoning and accuracy in educational, scientific, or assistive settings, it may also inadvertently enhance harmful uses, such as generating persuasive misinformation or biased content.

Nonetheless, we encourage future work to consider the broader implications of improved model steering and to pair performance improvements with safeguards and responsible deployment.

\smallskip
\section*{Acknowledgements}
This work was supported by a Tel Aviv University Center for AI and Data Science (TAD) grant and by Len Blavatnik and the Blavatnik Family foundation. This research was also supported by the Ministry of Innovation, Science \& Technology, Israel (1001576154) and the Michael J. Fox Foundation (MJFF-022407). SK is supported by the Google PhD Fellowship.

\bibliography{custom}

\appendix

\newpage

\section{Prompt Templates}
\label{sec:prompt_templates}

This section presents the prompt templates employed during various stages of the \textbf{ClusterFewshot} compilation process.

Tables~\ref{tab:gsm8k_oneshot_prompt}, \ref{tab:gsm8k_topk_fewshot_prompt}, and \ref{tab:gsm8k_diverse_fewshot_prompt} showcase representative templates used in the \textbf{GSM8K} task, covering both one-shot and few-shot prompting formats. These templates are adapted from the structured prompting scheme introduced in the original \textbf{BetterTogether} framework \cite{soylu-etal-2024-bettertogether}, preserving compatibility for meaningful comparison and reproducibility. Each template reflects a specific demonstration configuration used by our optimizer, and collectively they serve to illustrate the consistency and structure applied throughout the prompting pipeline.

\section{Semantic Clustering and Embedding Efficiency}
\label{sec:cluster_embedding_analysis}

ClusterFewshot uses external sentence encoders to structure the candidate demonstration pool before evaluation and sampling. We therefore analyze how different embedding models affect clustering quality and computational efficiency. {Detailed clustering-quality and embedding-efficiency results are provided in Table~\ref{tab:extended_eval_embed_clustering_gsm8k} and Table~\ref{tab:encoder_runtime_gsm8k}

{
Across four sentence encoders and three common clustering metrics
, Silhouette~\cite{rousseeuw1987silhouettes},
Calinski--Harabasz~\cite{calinskiharabasz}, and
Davies--Bouldin~\cite{daviesbouldin}, Silhouette and Calinski--Harabasz
consistently select $K=3$ and identify \texttt{gtr-t5-base} as the encoder
yielding the strongest cluster separation on \textbf{GSM8K}. In contrast,
Davies--Bouldin tends to favor larger $K$ values (8--10) and exhibits greater
variability across splits without corresponding downstream improvements.
}

Since embeddings are computed once per dataset split and reused throughout optimization, their computational cost directly affects overall pipeline efficiency. {Larger retrieval and LLM-derived encoders incur substantially higher runtime and memory overhead, whereas sentence-level encoders are significantly faster.} In particular, \texttt{all-mpnet-base-v2} and \texttt{gtr-t5-base} provide the best efficiency trade-off.
Overall, \texttt{gtr-t5-base} provides the best balance between clustering quality and computational efficiency. Based on these results, we adopt the Silhouette score with $K=3$ and the \texttt{gtr-t5-base} encoder for GSM8K in ClusterFewshot.

\section{Larger Models}

To examine whether the main efficiency trend extends to larger model settings,
Table~\ref{tab:gsm8k_qwen32b_standalone} reports standalone
prompt-optimization results across GSM8K, HotPotQA, and Iris using
\textbf{Qwen2.5-32B-Instruct} with AWQ quantization. Across all three
benchmarks, ClusterFewshot achieves the highest accuracy among the compared
methods while requiring the lowest compilation time among prompt optimizers.
This result strengthens the evidence that ClusterFewshot's efficiency advantage
holds at larger model scale, where repeated prompt compilation may become
increasingly costly, while preserving the accuracy benefits of optimized
few-shot prompting.

\section{Extended BetterTogether Results}
\label{sec:extended_results}

This section provides the extended experimental results referenced in the main paper.

Tables~\ref{tab:bt_qwen_gsm8k_all_runs}, \ref{tab:bt_qwen_hotpotqa_all_runs}, \ref{tab:bt_qwen_iris_all_runs}, \ref{tab:bt_llama_gsm8k_all_runs}, \ref{tab:bt_llama_hotpotqa_all_runs} and \ref{tab:bt_llama_iris_all_runs} report per-run performance scores for all \textbf{BetterTogether} strategies on the \textbf{GSM8K}, \textbf{HotPotQA}, and \textbf{Iris} benchmarks with Qwen2.5-7B-Instruct and Llama-3.2-3B-Instruct models, respectively. These results complement the aggregated scores presented in Table~\ref{tab:bt_all_models_qwen_llama} by including standard deviation estimates over three random seeds, providing a more comprehensive view of the stability and robustness of {all} prompt optimizers.

Notably, \textbf{ClusterFewshot} consistently exhibits lower variance across most strategies, benchmarks and models, underscoring its superior robustness relative to traditional bootstrap-based methods, \textbf{BFRS} {and \textbf{MIPROv2}}.

\section{Extended Ablation Study}
\label{sec:rebuttal_ablations}

This section provides extended ablation results supporting the analysis presented in the main paper. In addition to the performance summaries reported in Section~\ref{sec:sampling_retrieval_analysis}, we include full statistical results and additional experiments examining the design choices underlying ClusterFewshot. All experiments follow the same datasets and evaluation protocol described in Section~\ref{sec:experiments}.

Table~\ref{tab:individual_sampling_with_ci} reports the full statistical results for the individual cluster sampling strategies discussed in the main paper, situating them relative to full prompt optimizers. Similarly, Table~\ref{tab:retrieval_comparison_with_std} presents the corresponding statistical results for the comparison with \textbf{RetrievalFewshot (RFS)}.

Table~\ref{tab:zeroshot_gains} contextualizes the improvements of prompt optimizers relative to the zero-shot baseline, showing the absolute accuracy gains obtained by each method. Finally, Table~\ref{tab:oneshot_eval_sampling} reports results for alternative evaluation-set construction strategies used during one-shot scoring, comparing random validation subsets with cluster-central selections.

\section{Compute Resources}
\label{sec:compute_resources}

All experiments used to produce the results in Table~\ref{tab:bt_all_models_qwen_llama} were conducted on a single NVIDIA A100 GPU (80GB memory), across all tasks and three random seeds, with a total compute time of {\(\approx 160\)} GPU hours.

\begin{table*}[ht]
\centering
\small
\begin{tabular}{p{0.97\textwidth}}
\toprule
\textbf{Prompt (One-shot)} \\
\midrule
\textbf{Given a math word problem, generate the reasoning and final answer using Chain-of-Thought prompting.} \\

--- \\

\textbf{Role: system} \\
Your input fields are: \\
\hspace{1em}1. \texttt{question} (str) \\
Your output fields are: \\
\hspace{1em}1. \texttt{reasoning} (str) \\
\hspace{1em}2. \texttt{answer} (str) \\

All interactions will be structured in the following format: \\
\texttt{[[ \#\# question \#\# ]]} \\
\texttt{\{question\}} \\
\texttt{[[ \#\# reasoning \#\# ]]} \\
\texttt{\{reasoning\}} \\
\texttt{[[ \#\# answer \#\# ]]} \\
\texttt{\{answer\}} \\
\texttt{[[ \#\# completed \#\# ]]} \\

Objective: Given a question, produce a step-by-step reasoning and final answer. \\

--- \\

\textbf{Role: user (one-shot example)} \\

\texttt{[[ \#\# question \#\# ]]} \\
Annie goes to school. Today is her birthday, so Annie decided to buy some sweets for her colleagues. \\
Every classmate got 2 candies. In the end, Annie got left with 12 candies. \\
If there are 35 people in Annie's class in total, how much did Annie spend on candies, if one candy costs \$0.1? \\

\textbf{Role: assistant (completion)} \\

\texttt{[[ \#\# reasoning \#\# ]]} \\
Annie gave 2 candies to each of her 35 classmates: \(35 \times 2 = 70\). \\
She was left with 12 candies, meaning she originally had \(70 + 12 = 82\) candies. \\
Each candy costs \$0.1, so total cost is \(82 \times 0.1 = 8.2\) dollars. \\

\texttt{[[ \#\# answer \#\# ]]} \\
8.2 \\

\texttt{[[ \#\# completed \#\# ]]} \\

--- \\

\textbf{Role: user (actual question)} \\

\texttt{[[ \#\# question \#\# ]]} \\
Michelle is bringing sandwiches to work to give to her co-workers. \\
She gives 4 sandwiches to one of her co-workers and keeps twice this amount for herself. \\
If she had originally made 20 sandwiches, how many sandwiches does she have left to give to her other co-workers? \\

\textbf{(Model is expected to respond with reasoning and answer in same structure)} \\

\bottomrule
\end{tabular}
\caption{\textbf{ClusterFewshot} One-shot Prompting for \textbf{GSM8K} with \textbf{Chain-of-Thought} Reasoning, structured by Role}
\label{tab:gsm8k_oneshot_prompt}
\end{table*}

\begin{table*}[ht]
\centering
\small
\begin{tabular}{p{0.97\textwidth}}
\toprule
\textbf{Prompt (Few-shot, Top-3 Sampling)} \\
\midrule
\textbf{Given a math word problem, generate the reasoning and final answer using Chain-of-Thought prompting.}  \\

--- \\

\textbf{Role: system} \\
Your input fields are: \\
\hspace{1em}1. \texttt{question} (str) \\
Your output fields are: \\
\hspace{1em}1. \texttt{reasoning} (str) \\
\hspace{1em}2. \texttt{answer} (str) \\

All interactions will be structured in the following format: \\
\texttt{[[ \#\# question \#\# ]]} \\
\texttt{\{question\}} \\
\texttt{[[ \#\# reasoning \#\# ]]} \\
\texttt{\{reasoning\}} \\
\texttt{[[ \#\# answer \#\# ]]} \\
\texttt{\{answer\}} \\
\texttt{[[ \#\# completed \#\# ]]} \\

Objective: Given a question, produce step-by-step reasoning and a final answer. \\

--- \\

\textbf{Role: user (few-shot example 1)} \\
\texttt{[[ \#\# question \#\# ]]} \\
John's neighbor tells him to walk his dog for 1 hour each day for a total of \$10. \\
He does this for April, save for the 4 Sundays in April. \\
He later spent \$50 on books and gave his sister Kaylee the same amount. \\
How much money did John have left? \\

\textbf{Role: assistant (completion)} \\
\texttt{[[ \#\# reasoning \#\# ]]} \\
April has 30 days, and there are 4 Sundays, so John walks the dog for \(30 - 4 = 26\) days. \\
He earns \(26 \times \$10 = \$260\). \\
He spends \$50 on books and gives \$50 to his sister, totaling \$100 in expenses. \\
The remaining money is \(260 - 100 = 160\). \\

\texttt{[[ \#\# answer \#\# ]]} \\
\$160 \\
\texttt{[[ \#\# completed \#\# ]]} \\

--- \\

\textbf{Role: user (few-shot example 2)} \\
\texttt{[[ \#\# question \#\# ]]} \\
The ratio of boys to girls in a classroom is 3:5. \\
If there are 4 more girls than boys, how many students are in the classroom? \\

\textbf{Role: assistant (completion)} \\
\texttt{[[ \#\# reasoning \#\# ]]} \\
Let the number of boys be \(3x\), girls be \(5x\). \\
Since there are 4 more girls than boys: \\
\(5x = 3x + 4 \Rightarrow 2x = 4 \Rightarrow x = 2\). \\
So, boys = \(3 \times 2 = 6\), girls = \(5 \times 2 = 10\). \\
Total students = \(6 + 10 = 16\). \\

\texttt{[[ \#\# answer \#\# ]]} \\
16 \\
\texttt{[[ \#\# completed \#\# ]]} \\

--- \\

\textbf{Role: user (few-shot example 3)} \\
\texttt{[[ \#\# question \#\# ]]} \\
Lewis found 4 more items on the scavenger hunt than Samantha. \\
Samantha found four times as many items as Tanya, who found 4 items. \\
How many items did Lewis find? \\

\textbf{Role: assistant (completion)} \\
\texttt{[[ \#\# reasoning \#\# ]]} \\
Tanya found 4 items. Samantha found \(4 \times 4 = 16\) items. \\
Lewis found \(16 + 4 = 20\) items. \\

\texttt{[[ \#\# answer \#\# ]]} \\
20 \\
\texttt{[[ \#\# completed \#\# ]]} \\

--- \\

\textbf{Role: user (actual query)} \\
\texttt{[[ \#\# question \#\# ]]} \\
Michelle is bringing sandwiches to work to give to her co-workers. \\
She gives 4 sandwiches to one of her co-workers and keeps twice this amount for herself. \\
If she had originally made 20 sandwiches, how many sandwiches does she have left to give to her other co-workers? \\

\textbf{(Model is expected to respond with reasoning and answer)} \\

\bottomrule
\end{tabular}
\caption{\textbf{ClusterFewshot} Few-shot Prompting for \textbf{GSM8K} with \textbf{Chain-of-Thought} Reasoning using Top-3 sampled examples.}
\label{tab:gsm8k_topk_fewshot_prompt}
\end{table*}

\begin{table*}[ht]
\centering
\small
\begin{tabular}{p{0.97\textwidth}}
\toprule
\textbf{Prompt (Few-shot, Cluster Representative Sampling)} \\
\midrule
\textbf{Given a math word problem, generate the reasoning and final answer using Chain-of-Thought prompting.} \\

--- \\

\textbf{Role: system} \\
Your input fields are: \\
\hspace{1em}1. \texttt{question} (str) \\
Your output fields are: \\
\hspace{1em}1. \texttt{reasoning} (str) \\
\hspace{1em}2. \texttt{answer} (str) \\

All interactions will be structured in the following format: \\
\texttt{[[ \#\# question \#\# ]]} \\
\texttt{\{question\}} \\
\texttt{[[ \#\# reasoning \#\# ]]} \\
\texttt{\{reasoning\}} \\
\texttt{[[ \#\# answer \#\# ]]} \\
\texttt{\{answer\}} \\
\texttt{[[ \#\# completed \#\# ]]} \\

Objective: Given a question, produce step-by-step reasoning and a final answer. \\

--- \\

\textbf{Role: user (Cluster 1 Representative)} \\
\texttt{[[ \#\# question \#\# ]]} \\
There are three trees in the town square. The tallest tree is 150 feet tall. \\
The middle height tree is 2/3 the height of the tallest tree. \\
The shortest tree is half the size of the middle tree. How tall is the shortest tree? \\

\textbf{Role: assistant (completion)} \\
\texttt{[[ \#\# reasoning \#\# ]]} \\
The middle tree is \( \frac{2}{3} \times 150 = 100 \) feet tall. \\
The shortest tree is half of that: \( \frac{1}{2} \times 100 = 50 \) feet. \\

\texttt{[[ \#\# answer \#\# ]]} \\
50 \\
\texttt{[[ \#\# completed \#\# ]]} \\

--- \\

\textbf{Role: user (Cluster 2 Representative)} \\
\texttt{[[ \#\# question \#\# ]]} \\
Carl buys index cards for his class. He gives each student 10 index cards. \\
He teaches 6 periods a day and each class has 30 students. \\
If a 50-pack of index cards costs \$3, how much did he spend in total? \\

\textbf{Role: assistant (completion)} \\
\texttt{[[ \#\# reasoning \#\# ]]} \\
Carl teaches \(6 \times 30 = 180\) students. \\
Each student gets 10 index cards, so total needed = \(180 \times 10 = 1800\). \\
He needs \(1800 / 50 = 36\) packs. Each pack costs \$3, so total cost = \(36 \times 3 = \$108\). \\

\texttt{[[ \#\# answer \#\# ]]} \\
\$108 \\
\texttt{[[ \#\# completed \#\# ]]} \\

--- \\

\textbf{Role: user (Cluster 3 Representative)} \\
\texttt{[[ \#\# question \#\# ]]} \\
Kaylee needs to sell 33 boxes of biscuits. \\
She has sold 12 boxes of lemon, 5 of chocolate, and 4 of oatmeal biscuits. \\
How many more boxes does she need to sell? \\

\textbf{Role: assistant (completion)} \\
\texttt{[[ \#\# reasoning \#\# ]]} \\
Kaylee sold \(12 + 5 + 4 = 21\) boxes so far. \\
She needs to sell \(33 - 21 = 12\) more boxes. \\

\texttt{[[ \#\# answer \#\# ]]} \\
12 \\
\texttt{[[ \#\# completed \#\# ]]} \\

--- \\

\textbf{Role: user (actual query)} \\
\texttt{[[ \#\# question \#\# ]]} \\
Michelle is bringing sandwiches to work to give to her co-workers. \\
She gives 4 sandwiches to one of her co-workers and keeps twice this amount for herself. \\
If she had originally made 20 sandwiches, how many sandwiches does she have left to give to her other co-workers? \\

\textbf{(Model is expected to respond with reasoning and answer)} \\

\bottomrule
\end{tabular}
\caption{\textbf{ClusterFewshot} Few-shot Prompting for \textbf{GSM8K} using \textbf{Chain-of-Thought} Reasoning. Examples sampled from top-1 representatives of 3 semantic clusters, where the semantic structure was determined using the \texttt{gtr-t5-base} embedding model.}
\label{tab:gsm8k_diverse_fewshot_prompt}
\end{table*}

\begin{table*}[!htbp]
\centering
\small
\renewcommand{\arraystretch}{1.1}
\setlength{\tabcolsep}{7.5pt}
\begin{tabular}{l l c c c c}
\toprule
\textbf{Scoring Metric} & \textbf{Encoder} &
\textbf{Best K (train)} & \textbf{Score (train)} &
\textbf{Best K (dev)} & \textbf{Score (dev)} \\
\midrule
\multirow{4}{*}{\textbf{Silhouette} $\uparrow$}
  & Qwen3-Embedding-0.6B & 3 & 0.040 & \textemdash{} & \textemdash{} \\
\cmidrule(lr){2-6}
  & all-mpnet-base-v2     & 3 & 0.040 & \textemdash{} & \textemdash{} \\
\cmidrule(lr){2-6}
  & gtr-t5-base           & 3 & \textbf{0.057} & 3 & 0.051 \\
\cmidrule(lr){2-6}
  & bge-large-en-v1.5     & 3 & 0.043 & \textemdash{} & \textemdash{} \\
\midrule
\multirow{4}{*}{\textbf{Calinski--Harabasz} $\uparrow$}
  & Qwen3-Embedding-0.6B & 3 & 4.511 & \textemdash{} & \textemdash{} \\
\cmidrule(lr){2-6}
  & all-mpnet-base-v2     & 3 & 4.027 & \textemdash{} & \textemdash{} \\
\cmidrule(lr){2-6}
  & gtr-t5-base           & 3 & \textbf{5.559} & 3 & 12.49 \\
\cmidrule(lr){2-6}
  & bge-large-en-v1.5     & 3 & 4.231 & \textemdash{} & \textemdash{} \\
\midrule
\multirow{4}{*}{\textbf{Davies--Bouldin} $\downarrow$}
  & Qwen3-Embedding-0.6B & 10 & \textbf{2.744} & 8 & 4.106 \\
\cmidrule(lr){2-6}
  & all-mpnet-base-v2     & 9  & 2.760 & \textemdash{} & \textemdash{} \\
\cmidrule(lr){2-6}
  & gtr-t5-base           & 10 & 2.993 & \textemdash{} & \textemdash{} \\
\cmidrule(lr){2-6}
  & bge-large-en-v1.5     & 9  & 2.810 & \textemdash{} & \textemdash{} \\
\bottomrule
\end{tabular}
\caption{
\textbf{Selecting \(K\) and encoder via clustering-quality metrics \((K{=}2\text{–}10)\).}
We evaluate cluster quality on \textbf{GSM8K} across four sentence encoders using three metrics
(\emph{Silhouette} $\uparrow$, \emph{Calinski--Harabasz} $\uparrow$, \emph{Davies--Bouldin} $\downarrow$),
searching \(K \in \{2,\dots,10\}\).
\emph{Silhouette} and \emph{Calinski--Harabasz} consistently select \(K{=}3\) and agree that \texttt{gtr-t5-base} yields the strongest separation.
\emph{Davies--Bouldin} tends to prefer larger \(K\) (8–10) and shows less stability across splits without corresponding downstream gains.
Based on this study, we adopt \emph{Silhouette} with \(K{=}3\) and the \texttt{gtr-t5-base} encoder for GSM8K in ClusterFewshot.
}
\label{tab:extended_eval_embed_clustering_gsm8k}
\end{table*}

\begin{table*}[!htbp]
\centering
\small
\renewcommand{\arraystretch}{1.1}
\begin{tabular*}{\textwidth}{@{\extracolsep{\fill}}l c c c c@{}}
\toprule
\textbf{Encoder} &
\textbf{Dim} &
\textbf{Time (s)} $\downarrow$ &
\textbf{Ex/s} $\uparrow$ &
\textbf{Peak RSS Increase (GB)} $\downarrow$ \\
\midrule
Qwen3-Embedding-0.6B & 1024 & 23.48 & 3.8 & 4.04 \\
all-mpnet-base-v2    & 768  & 4.07  & 21.9 & 0.43 \\
\textbf{gtr-t5-base} & 768  & 4.28  & 20.8 & \textbf{0.13} \\
bge-large-en-v1.5    & 1024 & 13.42 & 6.6 & 1.37 \\
\bottomrule
\end{tabular*}
\caption{
\textbf{Embedding runtime and memory profiling on GSM8K (train split = 100 examples, CPU).}
We report embedding dimensionality, end-to-end embedding time, throughput, and peak resident set size (RSS) increase during embedding.
Results show substantial variation in computational cost across encoders, despite comparable embedding dimensionalities.
}
\label{tab:encoder_runtime_gsm8k}
\end{table*}


\begin{table*}[!htbp]
\centering
\small
\setlength{\tabcolsep}{4.5pt}
\renewcommand{\arraystretch}{1.05}
\begin{tabular}{lcccccc}
\toprule
\textbf{Optimizer} &
\textbf{GSM8K Acc.} & \textbf{Compile} &
\textbf{HotPotQA Acc.} & \textbf{Compile} &
\textbf{Iris Acc.} & \textbf{Compile} \\
&
\textbf{(\%)} & \textbf{(min)} &
\textbf{(\%)} & \textbf{(min)} &
\textbf{(\%)} & \textbf{(min)} \\
\midrule
Baseline & $86.72$ & -- & $20.40$ & -- & $70.00$ & -- \\
\textbf{ClusterFewshot} & $\mathbf{94.80}$ & $\mathbf{112.2}$ & $\mathbf{50.87}$ & $\mathbf{72.3}$ & $\mathbf{100.00}$ & $\mathbf{7.7}$ \\
BFRS & $94.46$ & $194.7$ & $49.07$ & $190.3$ & $94.00$ & $9.0$ \\
MIPROv2 & $93.40$ & $143.8$ & $50.13$ & $148.4$ & $96.00$ & $37.7$ \\
\bottomrule
\end{tabular}
\caption{
Standalone prompt-optimization results using Qwen2.5-32B-Instruct with AWQ quantization.
}
\label{tab:gsm8k_qwen32b_standalone}
\end{table*}

\begin{table*}[ht]
\centering
\small
\renewcommand{\arraystretch}{1.2}
\setlength{\tabcolsep}{8pt}
\begin{tabular}{l l c c c c}
\toprule
\textbf{Strategy} & \textbf{Prompt Optimizer} & \textbf{Run 1} & \textbf{Run 2} & \textbf{Run 3} & \textbf{Average (test)} \\
\midrule
\multirow{3}{*}{Prompts}                                
  & BFRS            & 86.57 & 85.36 & 84.84 & 85.59 $\pm$ 0.89 \\
  & MIPROv2         & 85.28 & 84.29 & 84.90 & 84.82 $\pm$ 0.50 \\
  & ClusterFewshot  & 88.85 & 86.87 & 87.63 & \textbf{87.78 $\pm$ 1.00} \\
\midrule
\multirow{3}{*}{Prompts $\rightarrow$ Prompts}           
  & BFRS            & 85.43 & 83.92 & 89.00 & 86.12 $\pm$ 2.61 \\
  & MIPROv2         & 82.93 & 82.02 & 83.46 & 82.80 $\pm$ 0.73 \\
  & ClusterFewshot  & 87.71 & 87.03 & 88.85 & \textbf{87.86 $\pm$ 0.92} \\
\midrule
\multirow{3}{*}{Weights $\rightarrow$ Prompts}           
  & BFRS            & 86.49 & 86.19 & 88.85 & 87.18 $\pm$ 1.46 \\
  & MIPROv2         & 81.87 & 83.99 & 83.31 & 83.06 $\pm$ 1.08 \\
  & ClusterFewshot  & 87.33 & 88.39 & 87.94 & \textbf{87.89 $\pm$ 0.53} \\
\midrule
\multirow{3}{*}{Prompts $\rightarrow$ Weights}           
  & BFRS            & 84.52 & 85.58 & 86.34 & 85.48 $\pm$ 0.91 \\
  & MIPROv2         & 74.43 & 86.72 & 77.54 & 79.56 $\pm$ 6.39 \\
  & ClusterFewshot  & 84.83 & 87.10 & 86.57 & \textbf{86.17 $\pm$ 1.19} \\
\midrule
\multirow{3}{*}{Prompts $\rightarrow$ Weights $\rightarrow$ Prompts} 
  & BFRS            & 85.05 & 79.51 & 81.71 & 82.09 $\pm$ 2.79 \\
  & MIPROv2         & 85.28 & 65.55 & 82.02 & 77.62 $\pm$ 10.58 \\
  & ClusterFewshot  & 82.85 & 89.15 & 86.04 & \textbf{86.01 $\pm$ 3.15} \\
\bottomrule
\end{tabular}
\caption{
Comparison of \textbf{BFRS}, \textbf{MIPROv2}, and \textbf{ClusterFewshot} as prompt optimizers across five \textbf{BetterTogether} strategies on the \textbf{GSM8K} benchmark using \textbf{Qwen2.5-7B-Instruct}. Each strategy was evaluated over three independent runs with different random seeds; we report the average accuracy on the held-out test set with its standard deviation. \textbf{MIPROv2} jointly optimizes instruction text and demonstrations, whereas \textbf{BFRS} and \textbf{ClusterFewshot} optimize demonstrations only. All optimizers compile few-shot contexts using up to 4 demonstrations per run.
}
\label{tab:bt_qwen_gsm8k_all_runs}
\end{table*}

\begin{table*}[ht]
\centering
\small
\renewcommand{\arraystretch}{1.2}
\setlength{\tabcolsep}{8pt}
\begin{tabular}{l l c c c c}
\toprule
\textbf{Strategy} & \textbf{Prompt Optimizer} & \textbf{Run 1} & \textbf{Run 2} & \textbf{Run 3} & \textbf{Average (test)} \\
\midrule
\multirow{3}{*}{Prompts}                                
  & BFRS            & 45.60 & 44.47 & 47.87 & 45.98 $\pm$ 1.73 \\
  & MIPROv2         & 46.60 & 43.40 & 44.80 & 44.93 $\pm$ 1.60 \\
  & ClusterFewshot  & 46.40 & 43.87 & 48.60 & \textbf{46.29 $\pm$ 2.37} \\
\midrule
\multirow{3}{*}{Prompts $\rightarrow$ Prompts}           
  & BFRS            & 42.13 & 46.27 & 42.73 & 43.71 $\pm$ 2.24 \\
  & MIPROv2         & 45.73 & 45.67 & 44.13 & 45.18 $\pm$ 0.91 \\
  & ClusterFewshot  & 44.33 & 47.33 & 44.33 & \textbf{45.33 $\pm$ 1.73} \\
\midrule
\multirow{3}{*}{Weights $\rightarrow$ Prompts}           
  & BFRS            & 43.67 & 40.07 & 43.67 & 42.47 $\pm$ 2.08 \\
  & MIPROv2         & 44.73 & 43.13 & 46.00 & \textbf{44.62 $\pm$ 1.44} \\
  & ClusterFewshot  & 41.33 & 44.33 & 42.87 & 42.84 $\pm$ 1.50 \\
\midrule
\multirow{3}{*}{Prompts $\rightarrow$ Weights}           
  & BFRS            & 42.47 & 44.53 & 45.40 & 44.13 $\pm$ 1.50 \\
  & MIPROv2         & 39.60 & 47.87 & 44.07 & 43.85 $\pm$ 4.14 \\
  & ClusterFewshot  & 42.93 & 46.00 & 43.73 & \textbf{44.22 $\pm$ 1.59} \\
\midrule
\multirow{3}{*}{Prompts $\rightarrow$ Weights $\rightarrow$ Prompts} 
  & BFRS            & 46.40 & 42.07 & 46.93 & 45.13 $\pm$ 2.67 \\
  & MIPROv2         & 42.73 & 44.33 & 43.00 & 43.35 $\pm$ 0.86 \\
  & ClusterFewshot  & 45.93 & 44.73 & 45.80 & \textbf{45.49 $\pm$ 0.66} \\
\bottomrule
\end{tabular}
\caption{
Comparison of \textbf{BFRS}, \textbf{MIPROv2}, and \textbf{ClusterFewshot} as prompt optimizers across five \textbf{BetterTogether} strategies on the \textbf{HotPotQA} benchmark using \textbf{Qwen2.5-7B-Instruct}. Each strategy was evaluated over three independent runs with different random seeds; we report the average accuracy on the held-out test set with its standard deviation. \textbf{MIPROv2} jointly optimizes instruction text and demonstrations, whereas \textbf{BFRS} and \textbf{ClusterFewshot} optimize demonstrations only. All optimizers compile few-shot contexts using up to 4 demonstrations per run.
}
\label{tab:bt_qwen_hotpotqa_all_runs}
\end{table*}

\begin{table*}[ht]
\centering
\small
\renewcommand{\arraystretch}{1.2}
\setlength{\tabcolsep}{8pt}
\begin{tabular}{l l c c c c}
\toprule
\textbf{Strategy} & \textbf{Prompt Optimizer} & \textbf{Run 1} & \textbf{Run 2} & \textbf{Run 3} & \textbf{Average (test)} \\
\midrule
\multirow{3}{*}{Prompts}
  & BFRS            & 84 & 74 & 76 & 78.00 $\pm$ 5.29 \\
  & MIPROv2         & 84 & 78 & 76 & 79.33 $\pm$ 4.16 \\
  & ClusterFewshot  & 86 & 82 & 82 & \textbf{83.33 $\pm$ 2.31} \\
\midrule
\multirow{3}{*}{Prompts $\rightarrow$ Prompts}
  & BFRS            & 94 & 74 & 80 & 82.67 $\pm$ 10.26 \\
  & MIPROv2         & 68 & 88 & 80 & 78.67 $\pm$ 10.07 \\
  & ClusterFewshot  & 82 & 92 & 86 & \textbf{86.67 $\pm$ 5.03} \\
\midrule
\multirow{3}{*}{Weights $\rightarrow$ Prompts}
  & BFRS            & 76 & 74 & 90 & 80.00 $\pm$ 8.72 \\
  & MIPROv2         & 82 & 86 & 84 & \textbf{84.00 $\pm$ 2.00} \\
  & ClusterFewshot  & 84 & 84 & 74 & 80.67 $\pm$ 5.77 \\
\midrule
\multirow{3}{*}{Prompts $\rightarrow$ Weights}
  & BFRS            & 82 & 82 & 88 & 84.00 $\pm$ 3.46 \\
  & MIPROv2         & 88 & 80 & 82 & 83.33 $\pm$ 4.16 \\
  & ClusterFewshot  & 82 & 90 & 84 & \textbf{85.33 $\pm$ 4.16} \\
\midrule
\multirow{3}{*}{Prompts $\rightarrow$ Weights $\rightarrow$ Prompts}
  & BFRS            & 88 & 86 & 80 & \textbf{84.67 $\pm$ 4.16} \\
  & MIPROv2         & 84 & 76 & 92 & 84.00 $\pm$ 8.00 \\
  & ClusterFewshot  & 86 & 76 & 84 & 82.00 $\pm$ 5.29 \\
\bottomrule
\end{tabular}
\caption{
Comparison of \textbf{BFRS}, \textbf{MIPROv2}, and \textbf{ClusterFewshot} as prompt optimizers across five \textbf{BetterTogether} strategies on the \textbf{Iris} benchmark using \textbf{Qwen2.5-7B-Instruct}. Each strategy was evaluated over three independent runs with different random seeds; we report the average accuracy on the held-out test set with its standard deviation. \textbf{MIPROv2} jointly optimizes instruction text and demonstrations, whereas \textbf{BFRS} and \textbf{ClusterFewshot} optimize demonstrations only. All optimizers compile few-shot contexts using up to 4 demonstrations per run.
}
\label{tab:bt_qwen_iris_all_runs}
\end{table*}

\begin{table*}[ht]
\centering
\small
\renewcommand{\arraystretch}{1.2}
\setlength{\tabcolsep}{8pt}
\begin{tabular}{l l c c c c}
\toprule
\textbf{Strategy} & \textbf{Prompt Optimizer} & \textbf{Run 1} & \textbf{Run 2} & \textbf{Run 3} & \textbf{Average (test)} \\
\midrule
\multirow{3}{*}{Prompts}
  & BFRS           & 75.49 & 79.74 & 79.51 & 78.24 $\pm$ 2.39 \\
  & MIPROv2        & 75.72 & 76.10 & 76.10 & 75.97 $\pm$ 0.22 \\
  & ClusterFewshot & 79.51 & 79.97 & 79.14 & \textbf{79.54 $\pm$ 0.42} \\
\midrule
\multirow{3}{*}{Prompts $\rightarrow$ Prompts}
  & BFRS           & 78.00 & 79.14 & 77.69 & 78.27 $\pm$ 0.76 \\
  & MIPROv2        & 77.54 & 77.62 & 77.01 & 77.39 $\pm$ 0.33 \\
  & ClusterFewshot & 80.88 & 80.80 & 81.49 & \textbf{81.05 $\pm$ 0.38} \\
\midrule
\multirow{3}{*}{Weights $\rightarrow$ Prompts}
  & BFRS           & 78.00 & 79.06 & 77.92 & 78.32 $\pm$ 0.64 \\
  & MIPROv2        & 75.27 & 76.18 & 74.51 & 75.32 $\pm$ 0.84 \\
  & ClusterFewshot & 79.82 & 80.12 & 80.58 & \textbf{80.17 $\pm$ 0.38} \\
\midrule
\multirow{3}{*}{Prompts $\rightarrow$ Weights}
  & BFRS           & 74.28 & 77.31 & 77.92 & 76.50 $\pm$ 1.95 \\
  & MIPROv2        & 75.42 & 73.29 & 76.71 & 75.14 $\pm$ 1.73 \\
  & ClusterFewshot & 76.93 & 77.92 & 78.15 & \textbf{77.66 $\pm$ 0.65} \\
\midrule
\multirow{3}{*}{Prompts $\rightarrow$ Weights $\rightarrow$ Prompts}
  & BFRS           & 79.89 & 76.40 & 78.98 & 78.42 $\pm$ 1.81 \\
  & MIPROv2        & 70.79 & 76.48 & 73.98 & 73.75 $\pm$ 2.85 \\
  & ClusterFewshot & 79.74 & 77.54 & 80.05 & \textbf{79.11 $\pm$ 1.37} \\
\bottomrule
\end{tabular}
\caption{
Comparison of \textbf{BFRS}, \textbf{MIPROv2}, and \textbf{ClusterFewshot} as prompt optimizers across five \textbf{BetterTogether} strategies on the \textbf{GSM8K} benchmark using \textbf{Llama-3.2-3B-Instruct}. Each strategy was evaluated over three independent runs with different random seeds, and we report the average accuracy on the held-out test set with its standard deviation. \textbf{MIPROv2} jointly optimizes instruction text and demonstrations, whereas \textbf{BFRS} and \textbf{ClusterFewshot} optimize demonstrations only. All optimizers compile few-shot contexts using up to 4 demonstrations per run.
}
\label{tab:bt_llama_gsm8k_all_runs}
\end{table*}

\begin{table*}[ht]
\centering
\small
\renewcommand{\arraystretch}{1.2}
\setlength{\tabcolsep}{8pt}
\begin{tabular}{l l c c c c}
\toprule
\textbf{Strategy} & \textbf{Prompt Optimizer} & \textbf{Run 1} & \textbf{Run 2} & \textbf{Run 3} & \textbf{Average (test)} \\
\midrule
\multirow{3}{*}{Prompts}
  & BFRS           & 37.20 & 35.60 & 38.07 & \textbf{36.96 $\pm$ 1.25} \\
  & MIPROv2        & 34.07 & 34.80 & 36.87 & 35.25 $\pm$ 1.45 \\
  & ClusterFewshot & 38.20 & 36.27 & 36.07 & 36.85 $\pm$ 1.18 \\
\midrule
\multirow{3}{*}{Prompts $\rightarrow$ Prompts}
  & BFRS           & 41.40 & 40.20 & 40.47 & 40.69 $\pm$ 0.63 \\
  & MIPROv2        & 32.87 & 36.13 & 35.47 & 34.82 $\pm$ 1.72 \\
  & ClusterFewshot & 40.20 & 40.93 & 41.67 & \textbf{40.93 $\pm$ 0.74} \\
\midrule
\multirow{3}{*}{Weights $\rightarrow$ Prompts}
  & BFRS           & 33.33 & 37.00 & 37.60 & 35.98 $\pm$ 2.31 \\
  & MIPROv2        & 32.73 & 32.33 & 30.20 & 31.75 $\pm$ 1.36 \\
  & ClusterFewshot & 36.13 & 37.60 & 38.33 & \textbf{37.35 $\pm$ 1.12} \\
\midrule
\multirow{3}{*}{Prompts $\rightarrow$ Weights}
  & BFRS           & 35.73 & 38.67 & 40.47 & \textbf{38.29 $\pm$ 2.39} \\
  & MIPROv2        & 28.00 & 33.40 & 27.93 & 29.78 $\pm$ 3.14 \\
  & ClusterFewshot & 36.40 & 39.20 & 34.87 & 36.82 $\pm$ 2.20 \\
\midrule
\multirow{3}{*}{Prompts $\rightarrow$ Weights $\rightarrow$ Prompts}
  & BFRS           & 39.00 & 32.53 & 37.27 & 36.27 $\pm$ 3.35 \\
  & MIPROv2        & 30.33 & 33.40 & 30.47 & 31.40 $\pm$ 1.73 \\
  & ClusterFewshot & 40.27 & 38.47 & 36.80 & \textbf{38.51 $\pm$ 1.74} \\
\bottomrule
\end{tabular}
\caption{
Comparison of \textbf{BFRS}, \textbf{MIPROv2}, and \textbf{ClusterFewshot} as prompt optimizers across five \textbf{BetterTogether} strategies on the \textbf{HotPotQA} benchmark using \textbf{Llama-3.2-3B-Instruct}. Each strategy was evaluated over three independent runs with different random seeds, and we report the average accuracy on the held-out test set with its standard deviation. \textbf{MIPROv2} jointly optimizes instruction text and demonstrations, whereas \textbf{BFRS} and \textbf{ClusterFewshot} optimize demonstrations only. All optimizers compile few-shot contexts using up to 4 demonstrations per run.
}
\label{tab:bt_llama_hotpotqa_all_runs}
\end{table*}

\begin{table*}[ht]
\centering
\small
\renewcommand{\arraystretch}{1.2}
\setlength{\tabcolsep}{8pt}
\begin{tabular}{l l c c c c}
\toprule
\textbf{Strategy} & \textbf{Prompt Optimizer} & \textbf{Run 1} & \textbf{Run 2} & \textbf{Run 3} & \textbf{Average (test)} \\
\midrule
\multirow{3}{*}{Prompts}
  & BFRS            & 66 & 74 & 68 & 69.33 $\pm$ 4.16 \\
  & MIPROv2         & 70 & 70 & 72 & 70.67 $\pm$ 1.15 \\
  & ClusterFewshot  & 70 & 66 & 78 & \textbf{71.33 $\pm$ 6.11} \\
\midrule
\multirow{3}{*}{Prompts $\rightarrow$ Prompts}
  & BFRS            & 62 & 72 & 70 & 68.00 $\pm$ 5.29 \\
  & MIPROv2         & 60 & 70 & 58 & 62.67 $\pm$ 6.43 \\
  & ClusterFewshot  & 78 & 66 & 82 & \textbf{75.33 $\pm$ 8.33} \\
\midrule
\multirow{3}{*}{Weights $\rightarrow$ Prompts}
  & BFRS            & 66 & 76 & 66 & 69.33 $\pm$ 5.77 \\
  & MIPROv2         & 64 & 68 & 50 & 60.67 $\pm$ 9.45 \\
  & ClusterFewshot  & 78 & 68 & 72 & \textbf{72.67 $\pm$ 5.03} \\
\midrule
\multirow{3}{*}{Prompts $\rightarrow$ Weights}
  & BFRS            & 70 & 60 & 76 & 68.67 $\pm$ 8.08 \\
  & MIPROv2         & 68 & 70 & 66 & 68.00 $\pm$ 2.00 \\
  & ClusterFewshot  & 72 & 70 & 72 & \textbf{71.33 $\pm$ 1.15} \\
\midrule
\multirow{3}{*}{Prompts $\rightarrow$ Weights $\rightarrow$ Prompts}
  & BFRS            & 64 & 72 & 64 & 66.67 $\pm$ 4.62 \\
  & MIPROv2         & 74 & 66 & 64 & 68.00 $\pm$ 5.29 \\
  & ClusterFewshot  & 70 & 78 & 78 & \textbf{75.33 $\pm$ 4.62} \\
\bottomrule
\end{tabular}
\caption{
Comparison of \textbf{BFRS}, \textbf{MIPROv2}, and \textbf{ClusterFewshot} as prompt optimizers across five \textbf{BetterTogether} strategies on the \textbf{Iris} benchmark using \textbf{Llama-3.2-3B-Instruct}. Each strategy was evaluated over three independent runs with different random seeds, and we report the average accuracy on the held-out test set with its standard deviation. \textbf{MIPROv2} jointly optimizes instruction text and demonstrations, whereas \textbf{BFRS} and \textbf{ClusterFewshot} optimize demonstrations only. All optimizers compile few-shot contexts using up to 4 demonstrations per run.
}
\label{tab:bt_llama_iris_all_runs}
\end{table*}

\newpage
\vspace{4pt}

\begin{table*}[t]
\centering
\footnotesize
\setlength{\tabcolsep}{4pt}
\renewcommand{\arraystretch}{1.05}
\begin{tabular}{lccc}
\toprule
\textbf{Method} & \textbf{GSM8K} & \textbf{HotPotQA} & \textbf{Iris} \\
\midrule
ClusterFewshot - Global Top-$k$  & $86.11 \pm 2.03$ & $42.20 \pm 1.83$ & $77.47 \pm 9.07$ \\
ClusterFewshot - Representatives & $85.98 \pm 1.45$ & $44.97 \pm 1.95$ & $86.27 \pm 6.04$ \\
ClusterFewshot - Cluster Random  & $85.24 \pm 3.19$ & $41.29 \pm 3.05$ & $76.00 \pm 7.65$ \\
ClusterFewshot - Centroids       & $86.39 \pm 2.37$ & $43.22 \pm 3.43$ & $79.60 \pm 8.31$ \\
\midrule
BFRS                              & $84.68 \pm 2.70$ & $44.48 \pm 3.30$ & $80.40 \pm 6.43$ \\
MIPROv2                           & $84.17 \pm 4.56$ & $44.21 \pm 2.64$ & $84.40 \pm 10.00$ \\
\midrule
\textbf{ClusterFewshot -- Hybrid} & $\mathbf{88.06 \pm 1.32}$ & $\mathbf{46.41 \pm 2.14}$ & $\mathbf{86.27 \pm 6.04}$ \\
\bottomrule
\end{tabular}
\caption{Performance of individual ClusterFewshot sampling strategies compared with full optimizers. {Reported values are Mean $\pm$95\% CI over 5 runs using Qwen2.5-7B-Instruct.}}
\label{tab:individual_sampling_with_ci}
\end{table*}

\vspace{-6pt}

\begin{table*}[t]
\centering
\footnotesize
\setlength{\tabcolsep}{4pt}
\renewcommand{\arraystretch}{1.05}
\begin{tabular}{lccc}
\toprule
\textbf{Method} & \textbf{GSM8K} & \textbf{HotPotQA} & \textbf{Iris} \\
\midrule
RFS-kNN                   & $84.70 \pm 0.54$ & $42.71 \pm 1.60$ & $76.67 \pm 7.57$ \\
RFS-MMR ($\lambda{=}0.2$) & $84.19 \pm 1.29$ & $43.38 \pm 0.53$ & $72.00 \pm 10.58$ \\
RFS-MMR ($\lambda{=}0.5$) & $83.18 \pm 0.61$ & $44.20 \pm 0.31$ & $76.67 \pm 10.07$ \\
RFS-MMR ($\lambda{=}0.8$) & $83.13 \pm 0.88$ & $43.20 \pm 0.41$ & $77.33 \pm 4.62$ \\
\midrule
\textbf{ClusterFewshot}   & $\mathbf{87.78 \pm 1.00}$ & $\mathbf{46.29 \pm 2.37}$ & $\mathbf{83.33 \pm 2.31}$ \\
\bottomrule
\end{tabular}
\caption{Comparison with retrieval-based few-shot baselines that select demonstrations at inference time using semantic similarity. {Reported values are Mean $\pm$ std over 3 runs using Qwen2.5-7B-Instruct.}}
\label{tab:retrieval_comparison_with_std}
\end{table*}

\begin{table*}[t]
\centering
\footnotesize
\setlength{\tabcolsep}{6pt}
\renewcommand{\arraystretch}{1.05}
\begin{tabular}{lccc}
\toprule
\textbf{Method} & \textbf{GSM8K} & \textbf{HotPotQA} & \textbf{Iris} \\
\midrule
Zero-shot & $82.78$ & $30.56$ & $42.67$ \\
\midrule
BFRS & $85.59\ (+2.81)$ & $45.98\ (+15.43)$ & $78.00\ (+35.33)$ \\
MIPROv2 & $84.82\ (+2.04)$ & $44.93\ (+14.38)$ & $79.33\ (+36.66)$ \\
\textbf{ClusterFewshot} & $\mathbf{87.78\ (+5.00)}$ & $\mathbf{46.29\ (+15.74)}$ & $\mathbf{83.33\ (+40.66)}$ \\
\bottomrule
\end{tabular}
\caption{Performance gains relative to the zero-shot baseline. All optimizers improve over the base model {Qwen2.5-7B-Instruct}, with ClusterFewshot yielding the largest gains across tasks.}
\label{tab:zeroshot_gains}
\end{table*}

\vspace{-6pt}

\begin{table*}[t]
\centering
\footnotesize
\setlength{\tabcolsep}{5pt}
\renewcommand{\arraystretch}{1.05}
\begin{tabular}{l cccc}
\toprule
\textbf{Sampling} & \textbf{\#Samples} & \textbf{GSM8K} & \textbf{HotPotQA} & \textbf{Iris} \\
\midrule
Random & 9  & $86.92 \pm 0.89$ & $45.18 \pm 1.39$ & $80.00 \pm 5.29$ \\
Random & 15 & $83.54 \pm 1.04$ & $43.58 \pm 4.12$ & $83.33 \pm 8.33$ \\
\textbf{Centrals} & \textbf{9} & $\mathbf{87.78 \pm 1.00}$ & $\mathbf{46.29 \pm 2.37}$ & $\mathbf{83.33 \pm 2.31}$ \\
Centrals & 15 & $87.33 \pm 0.30$ & $43.31 \pm 2.10$ & $82.67 \pm 3.06$ \\
\bottomrule
\end{tabular}
\caption{Effect of evaluation-set construction for one-shot scoring. Cluster-central validation subsets provide stable performance with moderate sample sizes, supporting the design choice used in ClusterFewshot. {Reported values are Mean $\pm$ std over 3 runs using Qwen2.5-7B-Instruct.}}
\label{tab:oneshot_eval_sampling}

\end{table*}

\end{document}